\documentclass[manuscript,screen]{acmart}

\usepackage{bm}
\usepackage[export]{adjustbox}
\usepackage{booktabs} 
\usepackage{multirow}

\usepackage[flushleft]{threeparttable}
 
\usepackage[symbol]{footmisc}

\AtBeginDocument{%
  \providecommand\BibTeX{{%
    \normalfont B\kern-0.5em{\scshape i\kern-0.25em b}\kern-0.8em\TeX}}}

\setcopyright{acmcopyright}
\copyrightyear{2018}
\acmYear{2018}
\acmDOI{XXXXXXX.XXXXXXX}

\acmConference[Conference acronym 'XX]{Make sure to enter the correct
  conference title from your rights confirmation emai}{June 03--05,
  2018}{Woodstock, NY}
\acmBooktitle{Woodstock '18: ACM Symposium on Neural Gaze Detection,
 June 03--05, 2018, Woodstock, NY} 
\acmPrice{15.00}
\acmISBN{978-1-4503-XXXX-X/18/06}

\begin{document}

\title{Conversation Group Detection With Spatio-Temporal Context}




\author{Stephanie Tan}
\affiliation{%
 \institution{Delft University of Technology}
 \country{Netherlands}}

\author{David M.J. Tax}
\affiliation{%
  \institution{Delft University of Technology}
  \country{Netherlands}}

\author{Hayley Hung}
\affiliation{%
  \institution{Delft University of Technology}
  \country{Netherlands}}




\begin{abstract}
  
  In this work, we propose an approach for detecting conversation groups in social scenarios like cocktail parties and networking events, from overhead camera recordings. We posit the detection of conversation groups as a learning problem that could benefit from leveraging the spatial context of the surroundings, and the inherent temporal context in interpersonal dynamics which is reflected in the temporal dynamics in human behavior signals, an aspect that has not been addressed in recent prior works. This motivates our approach which consists of a dynamic LSTM-based deep learning model that predicts continuous pairwise affinity values indicating how likely two people are in the same conversation group. These affinity values are also continuous in time, since relationships and group membership do not occur instantaneously, even though the ground truths of group membership are binary. Using the predicted affinity values, we apply a graph clustering method based on Dominant Set extraction to identify the conversation groups. We benchmark the proposed method against established methods on multiple social interaction datasets. Our results showed that the proposed method improves group detection performance in data that has more temporal granularity in conversation group labels. Additionally, we provide an analysis in the predicted affinity values in relation to the conversation group detection. Finally, we demonstrate the usability of the predicted affinity values in a forecasting framework to predict group membership for a given forecast horizon. 
  
\end{abstract}



\begin{teaserfigure}
\centering
  \includegraphics[width=\textwidth]{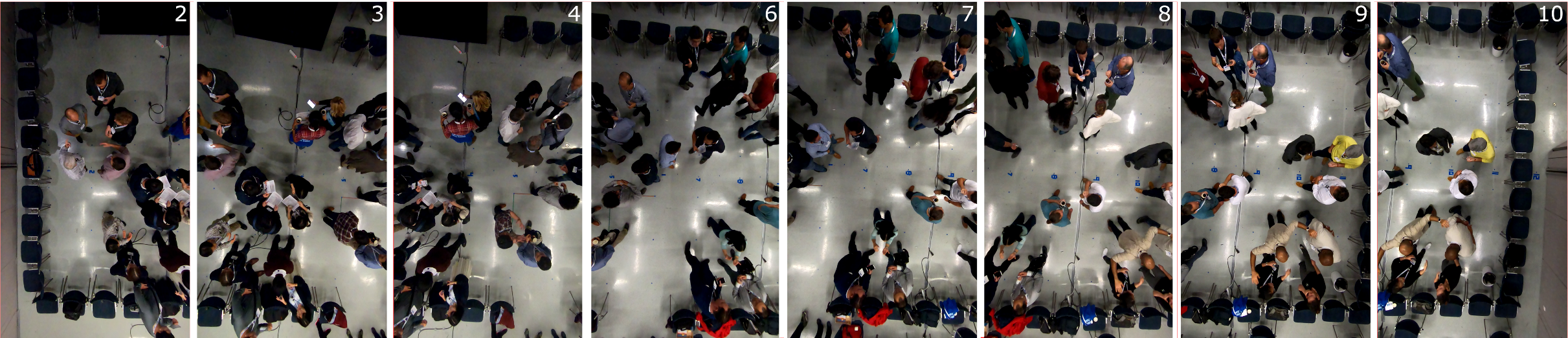}
  \caption{Snapshot of a typical social interaction area from the Conflab \cite{conflab} dataset. Faces blurred for privacy. }
  \label{fig:teaser}
\end{teaserfigure}


\maketitle

\section{Introduction}
\label{sec::introduction}
The automatic detection of conversation groups is an interesting problem for applications ranging from social surveillance \cite{cristani2013human,hung2011detecting,setti2015f} to social robotics \cite{bohus2014directions,vazquez2017towards,rios2015proxemics}. In social settings, such as cocktail parties or professional networking events, the floor for interactions at the venue consists of multiple conversation groups that dynamically adapt to the ebb and flow of the underlying human behaviors that determine the social interactions. Characterizing the interpersonal relationships that foster individuals to freely congregate to form a focused encounter and exchange information could help us understand more about interaction experience and quality \cite{kristoffersson2013towards, raj2020defining}. However, due to the complexity and subtleties in human social dynamics in changing environments which are also  context-specific, automatically detecting conversation groups in social scenes is an ongoing and challenging research topic. 

For this paper, we focus on identifying conversation groups (more specifically, free-standing conversation groups (FCGs) \cite{setti2015f}) in a social scene where individuals physically come together to interact with each other. In these settings they organize themselves into groups that define the physical partitions of who is interacting with whom. Figure \ref{fig:teaser} shows a representative scene for the types of social scenarios that we are interested in from an overhead viewpoint. People's use of physical space is known as \textit{proxemics} \cite{hall1966hidden}.The conversation groups can be of varying duration, size, and spatial arrangement. In practice, these conversation groups have often been conceptually formalized as \textit{facing-formation} (F-formations) \cite{hung2011detecting, setti2015f, alameda2015analyzing}. As introduced and defined by Kendon \cite{kendon1990conducting}, an F-formation is formed when two or more people arrange themselves onto a convex envelope to enclose an overlap of their interaction transactional segments (i.e., space in front of them where sight and hearing are most effective \cite{ciolek1980environment}). The interactants have equal and exclusive access to this overlap (i.e., \textit{o-space}). 



The challenges in formulating a social concept such as F-formations into a computational task for automated machine learning methods are two-fold: (i) representation of the scene and identifying the appropriate behavior cues to capture the underlying social dynamics, and (ii) the potential fuzziness that exists in the ground truth of group membership due to the fact that interpersonal relationships are not exactly binary in reality. 

A social scene such as Figure \ref{fig:teaser} may be represented by an interaction graph with nodes representing the individuals, and edges representing the relationship between two individuals. Conversation groups can be deduced from the information in this interaction graph, where individuals in the same conversation have greater edge weights (affinity) with one another, and vice versa. Each individuals have behavioral cues such as positions, head/body orientations, etc. that are shaped by the surroundings, and also change over time. Indeed, proximity with directional information could already be indicative of group membership if one simply considers 'who is standing with whom'. However, in social scenes such as Figure \ref{fig:teaser}, factors such as crowdedness and furniture layout define the spatial context that affects individual cues that determines conversation group membership. The temporal context in the behavioral cues  plays a role in the interactions which are dynamic in nature. One type of social dynamics that could be captured by movement cues (e.g., change in positions and orientations) is synchrony and mimicry patterns which are known to be important driver for affiliation and interpersonal rapport \cite{hove2009s,gleibs2016group,stel2010mimicking}. In terms of conversation dynamics, since multiple conversation floors are possible within one F-formation \cite{raman2019towards}, head orientations that change during turn-taking or other types of conversation dynamics are especially relevant to conversation schisms. The schisming phenomenon may lead to two or more distinct conversation groups \cite{egbert1997schisming} and hence possibly form new F-formations. Therefore, in this paper, we argue that even though F-formation in its original definition is a static concept, it is important to take into account the temporal context of the behavioral cues influencing the groups. 


The second fold of the challenge is the potential problem in how ground truths of group membership are defined from pairwise affinities. Past approaches of automatic detection of conversation groups have identified affinity and group memberships as binary and operationalized the task based on this design choice, albeit Zhang and Hung have investigated the subjectivity of annotating groups \cite{zhang2016beyond}. Although the assumption of the binary group membership which existing methods and ours hinge on, is valid to the extent of how they are reflected in the ground truth annotations, interpersonal relationships are not binary in reality. The temporally evolving affinity between two individuals, does not change from zero to one, or one to zero, instantaneously \cite{goffman1967interaction, goffman1955face, tuckman_developmental_1965}. Social interactions have a rite of passage, from greeting to leaving \cite{baehren2022saying}. We aim to understand how the affinity scores change in time and encapsulate the changing dynamics of the behavioral cues, and how they affect group detection, which are not apparent in hard assignments of group memberships. Our paper takes a step towards this direction which has not been the focus in previous works.




Following prior works \cite{swofford_dante_2020, Thompson2021}, our approach to conversation group detection consists of two stages: (1) we first estimate continuous pairwise affinities between all individuals in a social scene, and (2) we use an existing framework to cluster the individuals by leveraging a graph clustering based on Dominant Set (DS) to identify conversation groups \cite{hung2011detecting}. In order to account for the temporal context in the behavioral cues, we introduce a deep learning based Long Short-Term Memory (LSTM) network to predict pairwise affinity scores that determine the F-formation membership (annotated as ground truth of conversation groups in existing datasets). The inputs to our network are temporally aligned sequential inputs (including positions, and head and/or body orientation); the output of our network is the pairwise affinity value between one individual and \textit{all} other members in the social interaction scene. 

In contrast to existing works that output an affinity score of all pairs of individuals independently which scales quadratically with the number of people in the scene \cite{swofford_dante_2020, Thompson2021}, our design for person-wise output scales linearly and follows more intuitively from an egocentric application point of view (i.e. social robots). DS clustering is applied to the affinity matrix corresponding to each scene to detect the conversation groups. The clique formulation in DS is exploited to identify clusters in the interaction graph, which refines group detection since the affinity matrix (from estimated pairwise affinities) may not directly provide self-consistent and symmetric binary group memberships \cite{hung2011detecting,Hedayati2019,swofford_dante_2020}. 

Additionally, as opposed to previous approaches that only use the intermediate affinity scores as inputs for clustering for refined conversation group identification, we show the possibility in using the estimated pairwise affinity values from the past to forecast future affinity values, which could serve as an underpinning for understanding how conversation groups evolve. Even without formulating the problem explicitly as a future forecasting task, our model is able to anticipate affinity values due to the temporal continuity. Our main contributions are as follows:
\begin{itemize}
    \item We propose a novel LSTM-based affinity score model to approximate the likelihood of two people interacting in the same conversation group. The model includes a pooling module to account for the spatial context of social interactions, inspired by what \cite{Thompson2021,swofford_dante_2020} captured in their models. Using the proposed model that leverages (temporal) sequential input features in addition to a pooling module, we simultaneously account for the spatio-temporal context that affects pairwise affinities that determine the conversation groups. \item We provide an analysis over the predicted affinity scores in conversation group detection, characterized by Area under (receiver-operating) Curve (AUC) scores, followed by qualitative examples showing the continuity of affinity scores. We also show a comparison of affinity score processing (asymmetric vs. symmetric) for application in DS and the group detection performance with respect to scene dynamics.
    \item We demonstrate the usability of the predicted affinity values via a novel forecasting framework for affinity score prediction based on Gaussian Process Regression (GPR). The framework also provides inferential uncertainty quantification over the predictions of future conversation groups.  
\end{itemize}

\section{Related works}
Conversation group detection in situated interactions has been tackled by a variety of approaches stemming from different communities (computer vision \cite{Ricci_2015_ICCV, ricci2015uncovering, cristani2013human,setti2015f}, human-computer interaction \cite{Hedayati2019, Bohus2018ASI,Bohus2009, vazquez2017towards, Connolly2021}, etc.). This section discusses the representative works in this area. Conversation groups and the more formalized F-formation representation are analogous in terms of group detection in interaction scenes in past works.  

Many previous works, especially from the vision community, use features such as location and head/body orientations for the task of group detection \cite{alameda2015analyzing, ricci2015uncovering, hung2011detecting, vascon2016detecting,setti2015f}. These quantities could be obtained automatically from vision data using multi-camera surveillance setups (typically elevated side-views). 

Using these features, some methods for F-formation detection have been focused on optimization-based approaches to mathematically model the physical space. More concretely, a number of works hypothesised that the o-space can be generated from a noisy representation of the instantaneous view frustrum obtained from the head pose. Heat maps generated from samples projected from each individual's view frustrum were then used to identify o-spaces. Members of the F-formation were then re-identified as belonging to a particular o-space based a pre-defined metric of closeness. explicitly modeled the transactional segments of individuals, which then define the overlapping space that extends from the individuals (i.e. the o-space in F-formation definitions) \cite{setti2015f,gan2013temporal,Setti2013}. 

Another class of methods have formulated social scenes as an edge-weighted graph where each individual represents a node and the edge represents the pair-wise connection between individuals. These methods take the part of the F-formation definition related to equal mutual attention to synonymous with maximal cliques in edge weighted graphs. In early works, the pair-wise relationships were modeled using feature engineering based on location and orientation information \cite{vascon2016detecting,hung2011detecting, zhang2016beyond, zhang2018social}. Aggregating these estimated pairwise affinity values, the affinity matrix serves as inputs to graph clustering based on Dominant Set using game-theoretic approaches \cite{hung2011detecting,vascon2016detecting} to iteratively partition nodes to extract conversation groups. While the o-space is not explicitly modelled with these approaches, the maximal clique formulation implicitly models the o-space whilst also explicitly binding individuals to a specific group as part of the Dominant Set identification process. However, the representation of pairwise affinity, particularly when only location and orientation is used, forces a circular shape assumption to the F-formation that does not always happen in practice \cite{zhang2016beyond,zhang2018social}

To address this problem and enable more flexibility in modeling pairwise relationships given the surroundings, deep learning based approaches have been proposed. DANTE learns the affinity values by explicitly modeling dyadic and context interaction \cite{swofford_dante_2020} by using relative positions and head/body orientations after preprocessing as inputs to the model. More recently, \citet{Thompson2021} proposed a graph neural network (GNN) based approach that leverages the more general message-passing mechanism during training to predict affinities using raw signal data including absolute positions, accelerometer readings, and image. Similar to the preceding works \cite{hung2011detecting, vascon2016detecting, zhang2016beyond}, both of these deep learning based approaches also apply the learned affinities values inputs to DS graph clustering. \citet{schmuck2021growl} also proposed a GNN-based approach to predict interpersonal links, but as opposed to using DS graph clustering, a greedy agreement algorithm was applied to identify groups \cite{Hedayati2019}. 

Departing from using visually obtained features such as locations and orientations of individuals, some works have taken advantage of features from other modalities that have shown to be helpful when estimating pairwise affinities. For example, \citep{Thompson2021} take advantage of a combination of motion based features and visually obtained features. Gedik and Hung have shown that it is possible to estimate groups purely based on motion features as phenomena such as body movement synchrony in interactions are indicative of pairwise relationships \cite{gedik2018detecting}. However, in communication with the authors, the predicted affinities need to be significantly improved for before DS clustering would yield reasonable performance. This highlights that the nature of the problem lies between the modelling social dynamics and proxemics (i.e. positions and orientations). Approaches proposed by the ubiquitous and pervasive computing communities have relied on Bluetooth Smart (BLE) to measure proximity values in terms of Received Signal Strength Indicator (RSSI) values which capture distance and orientation information to some extent, represented by \cite{Marquardt2011}. In the case of \cite{Katevas2019}, data from motion sensors (accelerometer and gyroscope) were also incorporated with proximity features for multimodal detection of groups using smart phones. Other custom sensors have been developed to measure proximity, relative orientation, motion, and/or a combination thereof (e.g., light tags \cite{Montanari2018}, Rhythm badges \cite{Lederman2018} and the Midge \cite{midge}). These data also capture the useful information, such as direct measurement of closeness forming hypotheses of interactions already and the measurement of nuanced body motion, in determining conversation groups in social interactions, and methods developed based on these have the potential to scale more easily.

\begin{figure*}[ht!]
\includegraphics[scale=0.4]{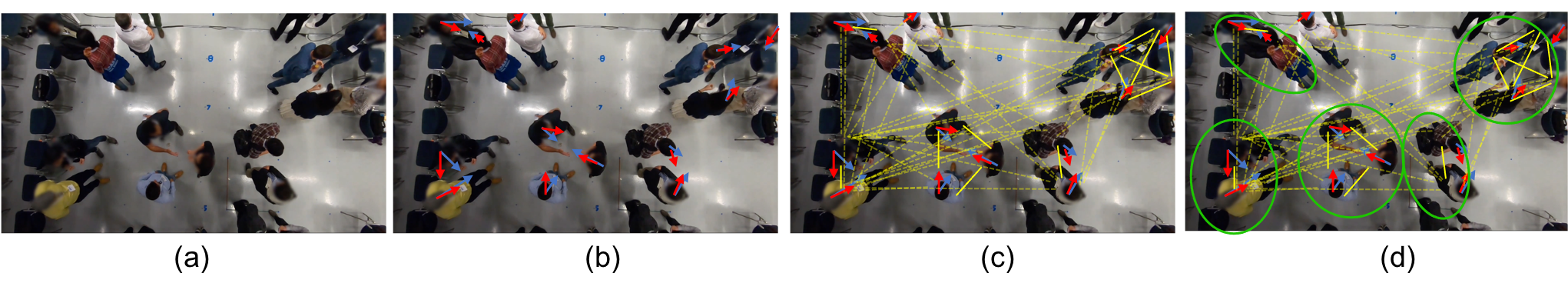}
\caption{Visualization of the derivation of the interaction graph: (a): the given scene from an overhead camera; (b): each subject has person-wise features, such as positions, orientations, etc.; (c): the edge weights (affinity scores) between each pair of subjects are predicted via the proposed neural network model; and (d): using the DS  \cite{hung2011detecting}, conversation groups are extracted as sets of subjects. 
}
\label{fig:workflow}
\end{figure*}

\section{Approach}
The overview of the approach to conversation group detection is illustrated in Figure \ref{fig:workflow}. Figure \ref{fig:workflow}(a) represents an example image from an interaction scene. The individual attributes such as positions, head and body orientations encode spatial information of an individual with respect to the scene (labeled in Figure \ref{fig:workflow}(b)).  

Module (b) represents the core of our contribution, which is a novel deep learning neural network for pairwise affinity estimation, based on a joint Long Short-Term Memory (LSTM) network that simultaneously accounts for the temporal context of the input signals and spatial context in the scene with spatially-motivated context pooling. In (c), the pairwise affinities are combined to a affinity matrix, and following previous approaches, Dominant Set was used to extract groups by iteratively identifying maximal cliques in edge-weighted graphs (module (d)).  Our contribution focuses on the neural network architecture for affinity prediction, and assumes that the inputs are acquired and preprocessed upstream. We use the Dominant Set clustering method on graphs downstream of affinity prediction because it is state-of-the-art method for this use-case. 

The details of the neural network architecture is described in Sec. \ref{sec::affinity}, and the details of the dominant set method is described in in Sec. \ref{sec::dominantset}.

\begin{figure*}[ht!]
\includegraphics[scale=0.4]{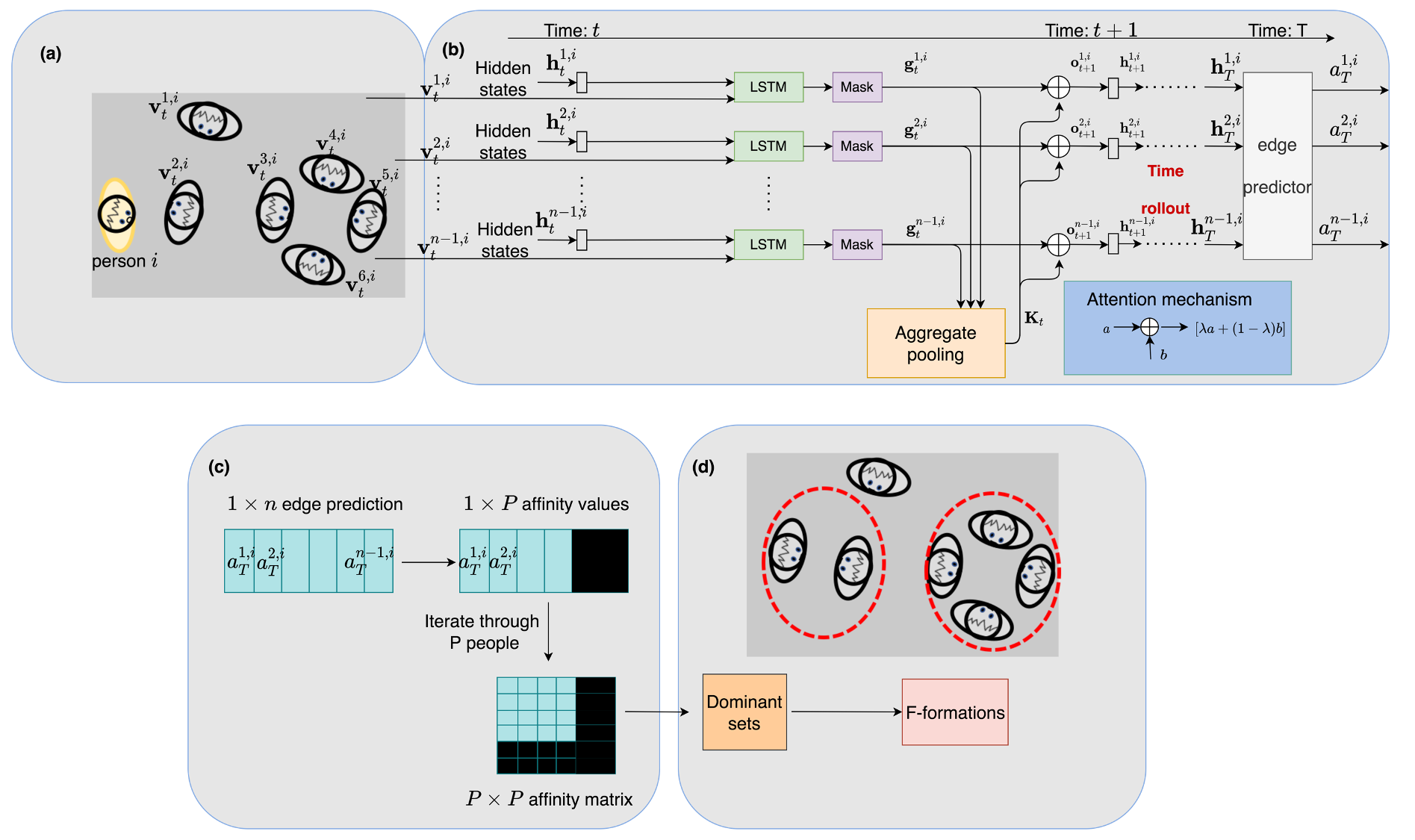}
\caption{Visualization of the methodology. (a): a scene representation of individuals and preprocessed features with respect to person i; (b): a graphical representation of the time roll out from $t$ to $t+1$ with the aggregate pooling layer along with the attention mechanism; (c) the filtering and aggregation of person-wise affinity output at $t=T$ into an affinity matrix; (d): application of DS clustering for group identification from the affinity matrix. }
\label{fig:arch}
\end{figure*}

\subsection{Affinity prediction}
\label{sec::affinity}
For a given social interaction scene $\mathcal{S}$, let $P_{t}$ represent the number of individuals in the scene at time step $t \in \{1,...,T\}$ where $T$ is the sequence length, and $n$ represent the maximum number of individuals present at all scenes of concern, i.e., $P_{t} \leq n$. Let $\bm{v}_{t}^{j,i} \in \mathbb{R}^N $ denote the $N$-dimensional feature vector for the $i^\text{th}$ member of the scene with respect to the $j^\text{th}$ member (with $1\leq i,j\leq P_t$ and $i\neq j$) at the sequence step $t$. The feature vector is a concatenation of features based on the following:
\begin{itemize}
    \item head and/or body orientations of member $i$,
    \item position, and head and/or body orientations of all members $j$ relative to the member $i$,
    \item indicator mask $I_t^j= \{1,0\}$ -- denoting if member $j$ is present in the scene at time  $t$ (assumed to be known a priori), 
\end{itemize}
the details of which are discussed in Section \ref{sec::implementation}.

Module (b) in Figure \ref{fig:arch} demonstrates one recurrent step of the proposed model from time $t$ to $t+1$. Let $\bm{h}_t^{j,i}\in\mathbb{R}^H$ denote the hidden representations associated with member $i$ at sequence step $t$, where $H$ is the dimension of hidden states (chosen as a hyperparameter). The hidden states at $t=0$ are initialized as $\bm{h}_0^{j,i} = \bm{0}$. To capture the spatial context defined by all members in the interaction scene, we pool the hidden states of all the members $j\neq i$ as follows. To discount for the persons absent in the scene at time $t$, the hidden state  are first processed through a masking layer (with element-wise multiplication) to obtain intermediate masked hidden states representation $\bm{g}_t^{j,i} = I^j_t \bm{h}_t^{j,i}$. Note that the  masked representation reflects the presence of members $j$ in the scene, which may be dynamically changing between different time steps. All masked representations $\bm{g}_t^{j,i}$ are processed through an \textit{aggregate pooling} layer to obtain a scene level representation $\mathcal{K}_t^i$ given by 
\begin{equation}
\label{poolingequation}
\displaystyle \mathcal{K}_t^i = \frac{\sum\limits_{j\neq i}\bm{g}_t^{j,i}}{\sum\limits_{j\neq i}I_t^{j}}.
\end{equation}
The context-pooled representation $\mathcal{K}_t^i$  is then combined with each of the individual hidden states $\bm{g}_t^{j,i}$ through an attention mechanism to obtain $\bm{o}_t^{j,i}$, 

\begin{equation}
\label{attentionequation}
\displaystyle \bm{o}_{t}^{j,i} = \lambda\mathcal{K}_t+(1-\lambda)\bm{g}_t^{j,i},
\end{equation}
where $\lambda\in[0,1]$ is a trainable parameter that adjusts the contributions from pairwise interaction and surrounding representations.

For each member $i$, the respective hidden state $\bm{h}_{t}^{j,i}$ as well as the concatenation of the feature $\bm{v}_{t+1}^{j,i}$ and the processed context representation $\bm{o}_t^{j,i}$ are passed through an LSTM cell $\mathcal{L}_{\tau}$ (parameterized by $\tau$) to obtain $\bm{h}_{t+1}^{j,i}$ (i.e., the hidden states for the subsequent time step), 
\begin{equation}
\label{eq:lstm}
	\bm{h}_{t+1}^{j,i} = \mathcal {L}_{\tau} ([\bm{v}_{t+1}^{j,i}; \bm{o}_{t}^{j,i}], \bm{h}_{t}^{j,i}).
\end{equation}

The LSTM operation $\mathcal L_\tau$ is described by the following series of transformations
	\begin{equation}
	\begin{aligned}
_{}	\bm{f}^{j,i}_{t+1} &= \sigma \left( \mathcal{W}_{\xi_f}\left([\bm{v}_{t+1}^{{j,i}}; \bm{o}^{j,i}_{t}; \bm{h}_{t}^{j,i} ] \right) \right) \qquad &&\text{(forget gate's activation vector)}\\
	\bm{i}^{j,i}_{t+1} &= \sigma \left( \mathcal{W}_{\xi_i}\left([\bm{v}_{t+1}^{{j,i}}; \bm{o}^{j,i}_{t}; \bm{h}_{t}^{j,i} ] \right) \right)\qquad &&\text{(input gate's activation vector)}\\
	\bm{o}^{j,i}_{t+1} &= \sigma \left( \mathcal{W}_{\xi_o}\left([\bm{v}_{t+1}^{{j,i}}; \bm{o}^{j,i}_{t}; \bm{h}_{t}^{j,i} ] \right) \right)\qquad &&\text{(output gate's activation vector)}\\
	\bm{\tilde c}^{j,i}_{t+1} &= \tanh \left( \mathcal{W}_{\xi_c}\left([\bm{v}_{t+1}^{{j,i}}; \bm{o}^{j,i}_{t}; \bm{h}_{t}^{j,i} ] \right) \right) \qquad &&\text{(cell input activation vector)}\\
	\bm{c}^{j,i}_{t+1} &= \bm{f}^{j,i}_{t+1} \odot \bm{c}^{j,i}_{t} + \bm{i}^{j,i}_{t+1} \odot \bm{\tilde c}^{j,i}_{t+1} \qquad &&\text{(cell state vector)}\\
	\bm{h}^{j,i}_{t+1} &= \bm{o}^{j,i}_{t+1} \odot \tanh\left(\bm{c}^{j,i}_{t+1} \right) \qquad &&\text{(output hidden state vector)},
	\end{aligned}
	\end{equation}
	where $\sigma$ is sigmoid activation, $\odot$ denotes the Hadamard product, and $\bm{c}^{j,i}_{t}$ and $\bm{c}^{j,i}_{t+1}$ denote the cell state at $t$ and $(t+1)$, respectively. $\mathcal{W}_{(\cdot)}$ denotes a  linear layer with  parameters indicated in the subscript. The trainable parameters are contained in the set $\tau=\{\xi_f, \xi_i, \xi_o, \xi_c\}$. Importantly, the LSTM parameters $\tau$ are shared among all the members in the scene.

After the time roll out in each LSTM time step until $t=T$, the hidden states $\bm{h}_{T}^{j,i}$ are passed through a linear layer parameterized by set of weights $\bm{W}_m$ and biases $\bm{b}_m$ to obtain the final pairwise edge predictions with respect to member $i$. Subsequently, they are passed through a sigmoid activation (denoted by $\sigma$) to obtain pairwise affinity $a^{j,i}\in[0,1]$ as
\begin{equation}
a^{j,i}=\sigma(\bm{W}_m \bm{h}_T^{j,i} + \bm{b}_m),
\end{equation}
where the values of 0 and 1 denote no and perfect pairwise affinity, respectively. The output of the model $a^{j,i}$ is strategically designed to be continuous, which lends naturally to a probability interpretation of pairwise interaction. We further motivate this choice, the performance evaluation, and the connection to downstream tasks such as conversation group forecasting in Section \ref{sec::metrics}.

To train the model, we use the mean squared error loss function given by
\begin{equation}
\label{lossfunction}
\ell= \sum_{\mathcal{S}}\sum_{j\neq i} \left( a^{j,i} - a_{GT}^{j,i} \right)^2,
\end{equation}
where $a_{GT}^{j,i}\in\{0,1\}$ represents the ground truth affinity value between member $j$ and member $i$. 

\subsection{Dominant set clustering}
\label{sec::dominantset}
As shown in Figure \ref{fig:arch} (c), for each member $i$ in the scene, the affinity prediction model predicts $n-1$ pairwise affinity values with respect to all other members $j$ for all time steps irrespective of whether they are visible at that moment or not. To evaluate the group identification performance, we use the output at the last time step at $t=T$. After filtering with the indicator masks, a $P \times P$ affinity matrix for the scene in question is obtained, where $P$ is the actual number of subjects at a particular scene.

After the predicted affinities are arranged into an affinity matrix, following prior approaches, \cite{hung2011,detecting,swofford_dante_2020, Thompson2021, vascon2016detecting}, the F-formations are extracted using Dominant Sets (DS) clustering. 
The resulting clusters representing F-formations could be of any size. The stopping criterion of the optimization formulation is either when the relative mutual affinity of internal nodes and external nodes of a dominant set do not satisfy the constraints, or when the mutual affinity of a group is lower than a chosen threshold. The second part of the stopping criterion enables improvement to detect singletons in the scene as it accounts for the global context (i.e., when there are only few people left after maximal clique extraction iterations, it is not likely that they are in the same group). We follow the implementation of F-formation clustering of \cite{swofford_dante_2020}. For the theoretical background and more detailed reference to the application of Dominant Set Clustering  for F-formation detection, please refer to \cite{bulo2017dominant,hung2011detecting}.

\section{Experimental setup}

\subsection{Baseline methods}
The baseline methods considered in this work are GTCG \cite{vascon2016detecting}, GCFF \cite{setti2015f}, and DANTE \cite{swofford_dante_2020}. GTCG and GCFF are both non deep learning based methods which rely on engineered position and orientation based features. GTCG models pairwise affinity values using distance between distributions over the plausible regions determined by the visual frustums, followed by a refined game theoretic approach for group extraction based on \cite{hung2011detecting} and \cite{Cristani2011}. GCFF models the probability of individuals belonging to o-space centers (i.e. center of conversation groups), and uses a graph-cut approach in conjunction with constraint based on direct access to extract the groups. 

 DANTE proposes a deep learning based approach to model pairwise affinities using positions and orientations, and utilizes the Dominant Set clustering for extracting conversation groups. During training of the deep learning model, DANTE uses data augmentation strategy   The recently proposed graph neural network based approach takes advantage of image-based features and a rich collection of social action semantic labels, in addition to proxemics and body motion based features, for conversation group detection \cite{Thompson2021}. We omit comparison against this recent approach for the scope of the paper, since the focus of our paper is on modeling the social dynamics in conversation scene using temporal information rather than a thorough investigation of using different input modalities.

\subsection{Datasets}
To align with the closest state-of-the-art approaches, we compare our method on the same representative datasets including Cocktail Party \cite{ZenEtAl2010} and SALSA \cite{AlamedaPineda2016}. We also report benchmark results on the recently released Conflab dataset \cite{conflab}, capturing professional networking social interactions in-the-wild. 

With conversation groups annotated at $1$Hz and behavioral cues sampled and annotated at $60$Hz then summarized to $1$Hz, the Conflab dataset is apt to investigate our research question which is leveraging temporal continuity in behavior cues and pairwise relationships in estimating affinity scores. In comparison, the conversation groups and behavior cues in Cocktail Party and SALSA are annotated at $1/5$ Hz and $1/3$Hz, respectively. We hypothesize that the temporal continuity in the signals and the ground truth of these two datasets can be decimated due to this sampling and annotation frequency. 

Most of the existing datasets were collected to serve F-formation detection using visual information, i.e. using an elevated side-view. Bounding boxes and head/body orientations are acquired either through automated methods or manual annotations. For datasets that have a top-down view, positions and orientations are acquired through manual annotations because automated methods result in error prone inputs to subsequent models \cite{conflab}. 
The overview of the datasets used is as follows:

\noindent -\textit{Cocktail Party} \cite{ZenEtAl2010}: contains 30 minute recordings of six people interacting with one another, captured by four elevated side view cameras in the corners of the space. Positions and head orientations of the subjects are obtained automatically using a particle filter-based body tracking method. The conversation groups were annotated at 1/5 Hz. 

\noindent -\textit{SALSA} \cite{AlamedaPineda2016}: contains 60 minute recordings of 18 people interacting with one another, captured by four elevated side view cameras in the corners of the space. Positions, head and body orientations, conversation groups of the subjects are annotated manually at 1/3 Hz. This dataset contains wearable sensor data captured by the Sociometric badges.

 \noindent -\textit{Conflab} \cite{conflab}: contains 15 minute recordings of 49 people interacting with one another, captured by 5 (non-overlapping) overhead cameras. Positions, head and body orientations of the subjects are annotated manually at 60Hz. Note that even though locations and orientations can be acquired automatically, some previous works have pointed out that the automatic methods produce erroneous results, especially in orientation estimation \cite{ZenEtAl2010}. To avoid confounding sources of errors in these behavioral cues which are quite nuanced as we motivated in Section \ref{sec::introduction}, we follow other works that have relied on provided ground truth data as inputs assuming that these will be provided upstream during application \cite{Tan2021}. Conversation groups are annotated manually at 1Hz.

\subsection{Evaluation metrics}
\label{sec::metrics}
We evaluate on both stages of our model: (1) pairwise affinity estimation and (2) group detection. For pairwise affinity prediction, the neural network is trained with binary ground truth, but the predicted affinities are continuous values between 0 and 1, which enables us to do further analysis using these affinity scores. Given the dynamic nature of the proposed model, we introduce an additional evaluation compared to the existing state-of-the-art methods. Existing methods (e.g., \cite{Thompson2021, swofford_dante_2020}) omit assessing the learned affinities only and use them directly for F-formation detection via DS clustering. We argue that there may be nuances in the learned pairwise continuous valued affinities that may anticipate changes in the group membership that may not be apparent from the hard cluster assignment.  

The evaluation metric for affinity estimation is the Area Under Curve (AUC) score of the Receiver Operator Curve (ROC). We use AUC due to the high imbalance of the data; there are typically far fewer positive pairwise memberships than negative in the entire scene. 


For the second stage of evaluating group detection, we used the standard evaluation metric used in prior work \cite{swofford_dante_2020,vascon2016detecting, hung2011detecting, zhang2016beyond, Cristani2011} which involves considering an entire group in the ground truth as a single sample. A detected group $k$ is considered to be correctly estimated if $\left\lceil{Thr \text{*} \|g_{k}\|}\right\rceil$ of the members are correctly estimated, where $\|g_{k}\|$ indicates the cardinality or the size of the ground truth group, and $\lceil x \rceil$ rounds $x$ to the next largest integer. The threshold $Thr\in [0,1]$ tunes the tolerance of the evaluation to the number of mis-attributed members in a group. It is commonly set to $Thr = \frac{2}{3}$ or $Thr = 1$, representing greater than majority overlap at 67\%  and complete overlap with the ground truth membership, respectively. A True positive (TP) is therefore any correctly detected group; false negative (FN) is a missed group; and a false positive (FP) is an estimated group that does not exist in the ground truth. The metrics for group detection performance is then computed using F1 measure over the entire image scene which could contain multiple groups. 

\subsection{Implementation details}
\label{sec::implementation}
In the case of Conflab, the features were extracted to align with the ground truth at 1Hz by averaging all 60 samples before the label. While it is desired to use a higher frequency signal, this preprocessing step should already allow capturing of social dynamics that exist on a second level, such as synchrony and convergence patterns \cite{kapcak2019estimating, quiros2021individual}, the associated postural sways \cite{funato2016smooth}, and some turn-taking dynamics (e.g., turn transitions). \cite{wilson2005oscillator}. 


The head and/or body orientation is given by the angular direction of the person’s body in $(-\pi, \pi]$. The relative positions of the group members are given by the radial distance (measured in camera or pixel coordinates) and angular orientation in $( -\pi, \pi]$. The circular mean of the body orientations of all the members in the scene are computed as a zero-degree reference for the scene. This addresses the discontinuity in angles as they wrap around $( -\pi, \pi]$ . All orientation related features are corrected by the same zero-degree reference. All features are normalized to $[0,1]$ via min-max scaling. Labels for conversation group membership are annotated manually and the annotation method are described in each dataset respectively. 

Similar to the experimental set up in DANTE, due to the small dataset size, all results are obtained by averaging the test splits using 5-fold cross validation.  The validation split is selected such that it separates the training set as much as possible from the test data in time. The test data of a fold is used for results whereas hyperparameters are selected based on validation data. The hyperparameters are hidden representation dimension of the LSTM and sequence length  of the input sequence. For the Conflab dataset, experimental results are obtained for all cameras (camera 2, 4, 6, and 8).

Since $P_t$ changes dynamically and the model is trained with a fixed size input using the maximum number of people in all scenes $n$, we pad the feature vector from $P_t$ to $n$ with dummy values of $-1$. As part of the feature vector, the indicator mask variable represents if a member is present at time $t$ such that the aggregate pooling layer does not account for the dummy subjects. 

\section{Results and Discussion}
\subsection{Overview}
To ensure a fairer comparison with existing methods, we use the same position and head and/or body orientation based feature set of the individuals in the scene for all methods. Table \ref{tab:overview} shows an overview of the results on baseline methods on the Cocktail Party, SALSA, and Conflab datasets. As we expected, the proposed method outperforms the baseline methods on the Conflab dataset because of the finer temporal granularity in the behavioral cues and group labels. As opposed to DANTE which excels in both Cocktail Party and SALSA dataset, the proposed method may not have leveraged the temporal context when the social dynamics is undersampled.




\begin{table}[h!]
\centering
\caption{F1 scores comparison $Thr=1$ across different methods. Standard deviation over test samples are shown in parenthesis. $^{*}$: results reported by \cite{swofford_dante_2020}; $^{\dagger}$: results reported by \cite{Thompson2021}}
\label{tab:overview}
\begin{tabular}{llll}
\hline
Method   & Cocktail Party & SALSA & Conflab \\ \hline
GTCG     & 0.29 (-) $^{*}$  & 0.44 (-) $^{*}$ & 0.40 (0.12)    \\
GCFF     & 0.64 (-) $^{*}$         & 0.41 (-) $^{*}$  & 0.31 (0.23)  \\
DANTE    & 0.58 (0.43) $^{\dagger}$   & 0.65 (-) $^{*}$  & 0.66 (0.35)   \\
Proposed & 0.48 (0.40)          & 0.46 (0.23)  & 0.73 (0.31)    \\ \hline
\end{tabular}

\end{table}


To assess whether or not the efficacy of the proposed model for the Conflab dataset is indeed associated to the frequency in conversation group labels, we subsample the Conflab dataset to $1/5$ Hz, to match that of the Cocktail Party dataset. The F1 performance at $Thr =1$ on this subsampled version of the Conflab dataset is 0.58 with standard deviation 0.32. So we see that even with the same setting, there is a decrease in performance due to an undersampling of key dynamic information that is leveraged by our proposed model. 

We observe that results of GTFF and GTCG decrease on the Conflab dataset compared to Cocktail Party and SALSA. This may be because the number of people in the scenes of Conflab are dynamically changing, as opposed to the fixed number of people in both Cocktail Party and SALSA (6 and 18 people, respectively), it may be harder to model o-space and determine overlapping transactional segments using a single parameter (stride), as participants' occupancy of floor space changes. DANTE still performs relatively well on Conflab as it also takes into account the spatial context of the surroundings. In addition to modeling the spatial context similar to DANTE, the proposed model relies on the sequential nature of the LSTM-based network to capture inherent temporal dynamics. 

With increased performance in affinity estimations (i.e., the model output), the performance in Dominant Set clustering for group extraction also improves. As we argue that the affinity estimations are critical not only because they are inputs to DS clustering, but also contain valuable information on how pairwise relationships change continuously over time, we include a more detailed analysis of the affinity values and their relationship with the DS clustering step in the next sections.

\subsection{Analysis of affinity values}
\label{subsec::affinity_analysis}

To uncover where the difference in group detection results in Table \ref{tab:overview} originates from, this section includes an analysis of where the proposed model differs from DANTE in terms of the predicted affinity value for test sets of the Cocktail Party and the Conflab dataset. Table \ref{tab::auc} shows a comparison of the predicted affinity value results from DANTE and the proposed method using the AUC metric. With its data augmentation strategy and benefiting from the pairwise output setup, the frame-based DANTE strategy works better for the sparsely sampled Cocktail Party dataset. For the Conflab dataset with the higher sampling frequency, the proposed approach takes into consideration the temporal continuity of labelled cues and affinity values with the data-efficient person-wise training and output, resulting in improved AUC scores that led to improved conversation group detection F1 scores. We posit that the temporal granularity could be too coarse in datasets such as the Cocktail Party for the proposed sequential model to be effective. 

\begin{table}[h!]
\caption{AUC results of DANTE and the proposed method for the Cocktail Party and Conflab dataset.}
\label{tab::auc}
\begin{tabular}{@{}lcc@{}}
\toprule
              & Cocktail Party & Conflab \\ \midrule
AUC (DANTE)&  0.92  &  0.91 \\                     
AUC (Proposed) & 0.83          & 0.93        \\ \bottomrule
\end{tabular}
\end{table}

\begin{figure*}[ht!]
\includegraphics[scale=0.20]{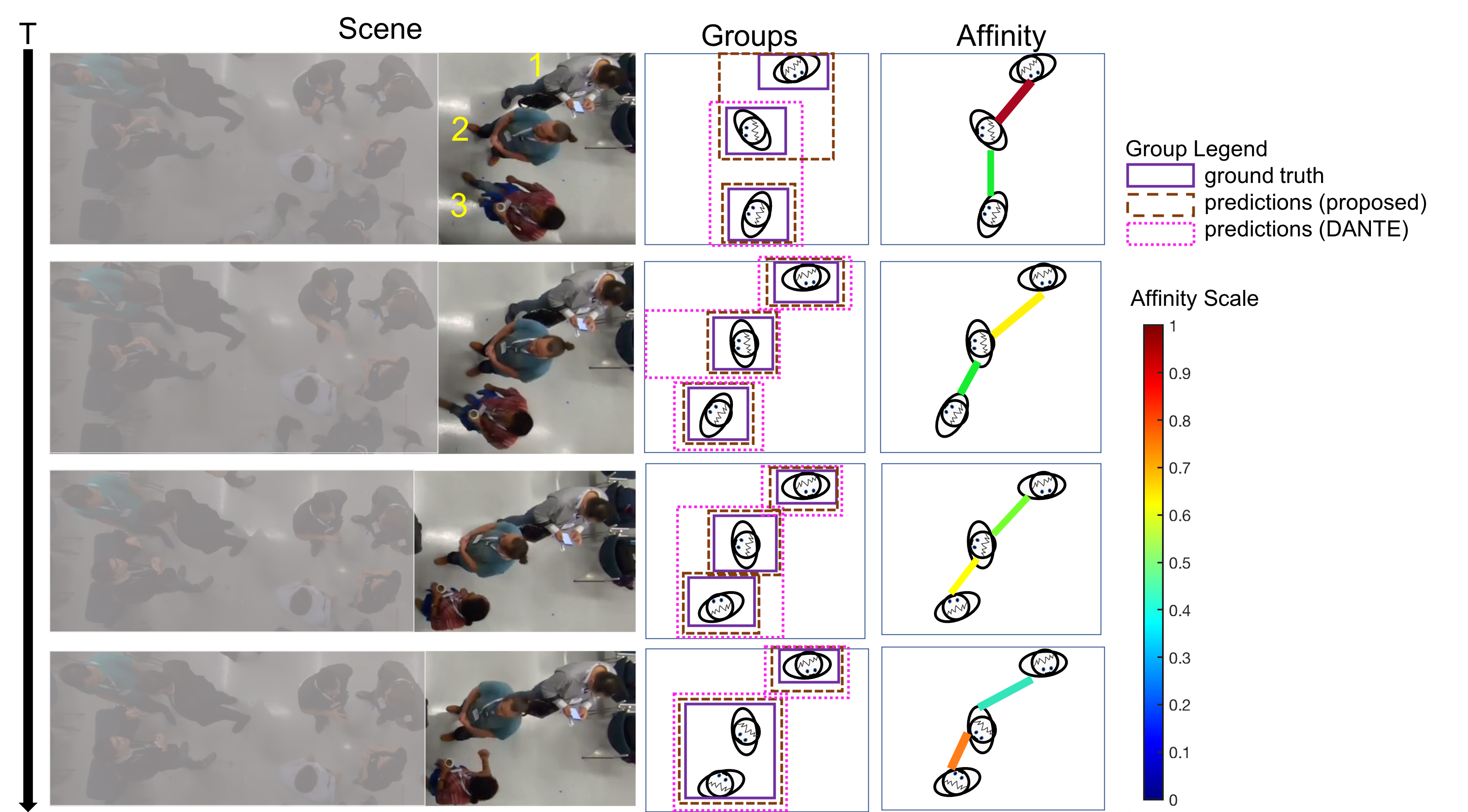}
\caption{Qualitative example of how affinity values change temporally, in relation to a newly formed conversation group.}
\label{fig:qualitative_affinity}
\end{figure*}

Figure \ref{fig:qualitative_affinity} shows a qualitative example of how the affinity values from the proposed model change temporally as a new conversation group (Subject 2 and 3) forms. We focus on the right side of the interaction floor in the sequence of the scenes shown. The groups provided by the ground truth, predictions from the proposed method, and DANTE are illustrated in the second column. The affinity scores between Subject 1 and 2, and Subjects 2 and 3 from the proposed method are visualized in the third column. The color and value correspondence is shown in the legend. The pairwise affinity scores between Subject 1 and 2 decrease over time, whereas the score between Subject 2 and 3 increase over time.  

\subsection{Affinity scores in Dominant Set clustering}
As the predicted affinity scores are continuous values, they are not perfect to directly extract group memberships. The pairwise values might differ and this results in discrepancy \cite{Hedayati2019, swofford_dante_2020}. Whether to symmetrize and how to symmetrize the predicted pairwise affinity scores is a design choice not thoroughly assessed previously. Options include using the asymmetric raw predicted affinity values, taking the minimum, average, or maximum of the pairwise affinity values. The (a)symmetry could be illustrative of the individual's intention in interacting with the other person, and affect the group clustering performance as this factor may have also affected how the annotators perceived group memberships. 

In Table \ref{tab::symmetry_aff}, we show the sensitivity F1 scores of symmetrizing the affinity matrix using different strategies for the Conflab dataset. The results show that averaging the pairwise affinities leads to improved F1 score in group detection at $Thr=1$. This implies that while the asymmetric values are interesting in differentiating people's likelihood in interacting with each other, a symmetric and averaged representation is better aligned with the binary group membership.

\begin{table}[!htb]
\begin{minipage}{.42\linewidth}
    \caption{F1 scores comparison ($Thr=1$) for the Conflab datset between different strategies of processing the affinity scores as inputs to DS clustering. Standard deviation over test samples are shwon in parenthesis.}
    \centering
    \label{tab::symmetry_aff}
    \begin{adjustbox}{max width=\linewidth}
    \begin{tabular}{@{}lcccc@{}}
\toprule
   & raw  & average & minimum & maximum \\ \midrule
F1 & 0.69 (0.33) & 0.73 (0.31)    & 0.72 (0.30)    & 0.65 (0.35)   \\ \bottomrule
\end{tabular}
    \end{adjustbox}
\end{minipage}\hfill
\begin{minipage}{.55\linewidth}

\caption{F1 scores comparison ($Thr$=1) for the Conflab dataset with respect to scene dynamics quantified by $D$. Standard deviation over test samples are shown in parenthesis.} 

\label{tab::dynamics_table}
\centering
\begin{adjustbox}{max width=\linewidth}
\begin{tabular}{@{}llllll@{}}
\toprule
         & $D=1$          & $D=2$          & $D=3$          & $D=4$          & $D\geq5$ \\ \midrule
DANTE    & 0.69 (0.36)  & 0.65 (0.36)  & 0.78 (0.27)  & 0.68 (0.32)  & 0.66 (0.23)       \\
Proposed & 0.71 (0.34)  & 0.69 (0.34)  & 0.77 (0.31)  & 0.73 (0.33)  & 0.77 (0.32)       \\
Delta    & 0.028 (0.33) & 0.037 (0.32) & -0.004 (0.27) & 0.039 (0.33) & 0.1 (0.35)       \\ \bottomrule
\end{tabular}
\end{adjustbox}
\end{minipage}
\end{table}

\subsection{Performance with respect to scene dynamics}
We highlight the efficacy of the proposed method when estimating groups especially in scenes that contain more instances of group formations, breaking, and reforming. These events quantify scene dynamics as they imply changes in one or more conversation group reorganization. We define these events based on group presence in the past and future (i.e. a new group is formed if it doesn't exist before; a group is broken if it doesn't sustain to the next time step; and a group is reformed when it exists but breaks in the past and now the same members reunite). 

We characterize the scenes in the Conflab dataset using this measure of scene dynamics. In Table \ref{tab::dynamics_table}, we showcase the performance difference (indicated by Delta) of group detection performance using F1 scores at $Thr = 1$ between DANTE and the proposed method. Delta is calculated by the F1 scores obtained from the proposed method substracted by that of DANTE. We show the results at different level of scene dynamics denoted as $D$, where $D$ is the sum of all instances of group formations, breaking, and reformation. We observe a slight upward trend of the proposed method's improved performance (i.e., Delta) as scenes become more dynamic (with the exception of $D$=3). When $D \geq 5$ (corresponding to high scene dynamics, whereas $D=0$ for most of the time where groups are stable), the advantage of the proposed method over DANTE on average is at 0.1. This further shows that the temporal context before a group event takes place may be beneficial in estimating conversation groups.




\subsection{Conversation group forecasting}
Using the temporal context of pairwise affinity scores, we further introduce a conversation group forecasting framework. Given a sequence of edge weights $a^{\text{edge}}_{t}$ where $t = 0,1,....,T$, which is the averaged value between each pair of individuals $i$ and $j$ (i.e., $\frac{a_{i,j}+a_{j,i}}{2}$), we predict  $a^{\text{edge}}_{t+1}, a^{\text{edge}}_{t+2},...,a^{\text{edge}}_{t+Z}$ where $Z$ is the time forecast horizon. For each $a^{\text{edge}}_{t}$ sequence, we fit a Gaussian Process Regressor (GPR) to provide uncertainty measure over the predictions.

GPR assumes a kernel that determines the covariance over target functions and uses the observations to obtain a likelihood function. A new posterior distribution can be computed based on Bayes' theorem. The choice of kernel characterized by a covariance function that measures the similarity between data points is an essential component in GPR. For the purpose of this study, this covariance function is chosen to be the popular Radial Basis Function (RBF). For more technical background on GPR, please refer to \cite{rasmussen2003gaussian}. 

For each GPR corresponding to an edge, we use the observed samples $A_{\text{edge,t}}$ to optimize the length-scale hyperparameter in the RBF kernel based on maximum-log-likelihood estimation. Using the fitting regressor function, a set of posterior samples up to the maximal forecast time horizon are predicted. Leveraging the probabilistic nature of GPR, we evaluates $N$ samples drawn from the GPR at given inputs (in our case, a time step in the future). These $N$ samples drawn from the Gaussian distribution at given $t$ provide a range of values for the edge weight, and ultimately result in an uncertainty quantification of group membership (after aggregating edge forecasts into affinity matrix and application of the DS clustering  $N$ times). 

\begin{table}[ht!]
\caption{Performance of forecasting conversation groups at different future time steps for the Conflab dataset (cam 6). Note that the column $t =T$ indicates the detection results. Uncertainty quantified by standard deviation across all scenes are shown in parenthesis.}
\label{tab::forecasting}
\begin{tabular}{@{}lllllll@{}}
\toprule
               & t = T & t = T+1 & t = T+2 & t = T+3 & t = T+5 & t =T +10\\ \midrule
F1 @ $Thr= 2/3$  & 0.90  &   0.88 (0.03)     &   0.86 (0.04)      & 0.84 (0.05)  & 0.80 (0.07) & 0.76 (0.08)      \\
F1 @ $Thr = 1$   & 0.76  &  0.73 (0.06)      & 0.69 (0.06)       &  0.69 (0.08)   & 0.66 (0.10) & 0.62 (0.11)   \\ \bottomrule
\end{tabular}
\end{table}

 From the validation sets of the data, we found that a sequence length of 10 was the optimal hyperparameter for affinity prediction and hence, we set $T = 10$ to acquire corresponding observed samples of affinity scores to fit the GPR for this forecasting task. The forecast horizon $Z$ represents the time steps beyond $T$ (measured in seconds for the Conflab dataset). Note that the fitted GPR could be sampled continuously; we chose a set of discrete time steps beyond $T$ for the scope of this paper. Table \ref{tab::forecasting} shows the forecasting results of predicting the conversation groups in Conflab (cam 6) using the aforementioned approach. We report the averaged F1 scores from evaluating the $N$ affinity matrix instances of each scene (i.e. aggregated from using the $N$ samples of affinities drawn from each edge) for all scenes. The results show that there is a decreasing trend in the group detection performance as the forecast horizon extends, while the uncertainty in the group prediction in future scenes increases. 
 

\section{Conclusion and Future works}
In this work, we introduce and evaluate a deep learning joint-LSTM based neural network for pairwise affinity prediction, followed by DS clustering approach, for the task of conversation group detection in social settings such as cocktail parties and networking events. We motivate this LSTM-based approach to leverage the inherent temporal dynamics of human behaviors who could affect interactions and conversation groups. We showed that for the Conflab dataset (which has more temporal granularity compared to other existing datasets), our method shows improved performance in pairwise affinity predictions and therefore, leading to improved performance of conversation group detection. We further showed an analysis of the predicted affinity predictions and how they change overtime, which could be indicative of moments leading up to group formations and breaking. Lastly, we provide a forecasting framework based on our approach which predicts conversation groups at future time steps.

One of the limitations of this work include its performance in sparsely labelled data, such as for the Cocktail Party and SALSA dataset. The lower annotation frequency implies more varied conversation groups between time steps, and that the continuously changing group behavior in real life is not captured in the ground truth. Moreover, our use of ground truth features was partially motivated by what was provided in the existing datasets, and allows us to investigate the model performance without potential confounding sources of errors. However, this choice also does not shed light on the sensitivity of the performance with automatically acquired features, which would ultimately be more relevant in automatic systems (e.g., a social robot). 

The model architecture could be further revised to take advantage of multimodal data at full sampling rates, for video, audio, and body movement motion. We note that the Conflab dataset contains manually annotated positions and orientations at 60Hz, as well as full 9 Degrees-of-Freedom IMU motion data captured at 56Hz sampling rate and speaking status annotations at 60Hz. While the trade-off among the difficulty of acquiring all of these data in an application setting, building and deploying a larger model, and the potential increase in performance should be considered, we believe that using more expressive modalities at finer temporal resolution, conversational group dynamics may be more thoroughly captured. 

For further extension, the proposed forecasting framework presents an opportunity for researchers to detect  individuals' intent to interact with others. More socially intelligent automated systems can be built if they are able to forecast affinities as a proxy for intention. Based on whether or not the predictions align with what actually occurs in the future, applications that are cognizant of what humans \textit{plan} or \textit{want} to do can be designed to enable better social interactions. 

\begin{acks}

\end{acks}

\bibliographystyle{ACM-Reference-Format}
\bibliography{ff}










\end{document}